\documentclass{article}

\usepackage[nonatbib]{neurips_2019}
\usepackage[utf8]{inputenc}
\usepackage[T1]{fontenc}
\usepackage{url}
\usepackage{booktabs}
\usepackage{amsfonts}
\usepackage{nicefrac}
\usepackage{microtype}
\usepackage{graphicx}
\usepackage{xcolor}
\usepackage{lipsum}
\usepackage{amsmath}
\usepackage{geometry}

\usepackage{booktabs}
\usepackage[style=numeric,backend=biber]{biblatex}
\bibliography{references}

\title{
  Exploring intra-task relations to improve meta-learning algorithms \\
  \vspace{1em}
}
\author{
  Prabhat Agarwal \\
  Stanford University \\
  \texttt{prabhat8@stanford.edu} \\
  \And
   Shreya Singh\\
   Stanford University \\
  \texttt{ssingh16@stanford.edu} \\
}

\begin{document}

\maketitle
\section{Abstract}

Meta-learning has emerged as an effective methodology to model several real-world tasks and problems due to its extraordinary effectiveness in the low-data regime. There are many scenarios ranging from the classification of rare diseases to language modelling of uncommon languages where the availability of large datasets is rare. Similarly, for more broader scenarios like self-driving, an autonomous vehicle needs to be trained to handle every situation well. This requires training the ML model on a variety of tasks with good quality data. But often times, we find that the data distribution across various tasks is skewed, i.e.the data follows a long-tail distribution. This leads to the model performing well on some tasks and not performing so well on others leading to model robustness issues. Meta-learning has recently emerged as a potential learning paradigm which can effectively learn from one task and generalize that learning to unseen tasks. 

However, it is often difficult to train a meta-learning model due to stability issues. Negative transfer (cite: to transfer on not to), which is commonly seen in transfer learning, is one of the main reasons for this instability. Akin to transfer learning where negative transfer can actually hinder performance if the tasks are too dissimilar, understudied effects of different task interactions can affect the performance in meta-learning as well. It will be useful to study the task distribution of meta-train and meta-test tasks and leverage any external source of information about these tasks which can help us create more informed mini-batches instead of the status-quo of randomly selecting tasks for the mini-batch.

In this study, we aim to exploit external knowledge of task relations to improve training stability via effective mini-batching of tasks. We hypothesize that selecting a diverse set of tasks in a mini-batch will lead to a better estimate of the full gradient and hence will lead to a reduction of noise in training. 

Our contributions are two-fold in this project: Firstly, we leverage WordNet to build the class-relation graph for the 100 classes of the mini-Imagenet dataset. We then generate clusters in that graph based on the node(class) distances. After the generation of class clusters, we effectively sample classes from them and generate tasks of varying level of complexities. Later, we also generate an artificial dataset by using a backward approach. Specifically, we first take the WordNet hierarchy and select 15 clusters (subtrees). We then sample 10 classes from each of these clusters to get 150 classes in total. Then for each of these classes, we again sample 100 images from the ImageNet dataset to generate an artificial dataset having the prior of WordNet class clusters. One thing to note is that we use these clusters not only to generate tasks of varying levels of complexity for the meta-training phase, but, we also use the clustering information to sample tasks for the meta-testing phase as well. This helps us to evaluate the dependency of meta-test performance over the meta-train performance for varying level of task complexity distribution.

Secondly, we test our hypothesis through training two meta learners - MAML and ProtoNet over the new task distributions and study the correlation between various combinations of task complexity distributions of the meta-train and meta-test phase. We also hypothesize that constructing meta-train tasks such that it is not very different from the meta-test tasks will reduce the effects of negative transfer and potentially lead to faster convergence. This can also improve model performance via efficient meta-train task selection.

\section{Introduction} 

Meta-Learning is one of the fastest-growing areas of research in the field of Machine Learning. Meta-learning, in the machine learning context, is the use of machine learning algorithms to assist in the training and optimization of other machine learning models. The general idea of meta-learning is 'learning how to learn'. Meta-learning is the ability of an artificially intelligent machine to learn how to carry out various complex tasks, taking the principles it used to learn one task and applying it to other tasks. 

\begin{figure}[ht]
    \centering
    \includegraphics{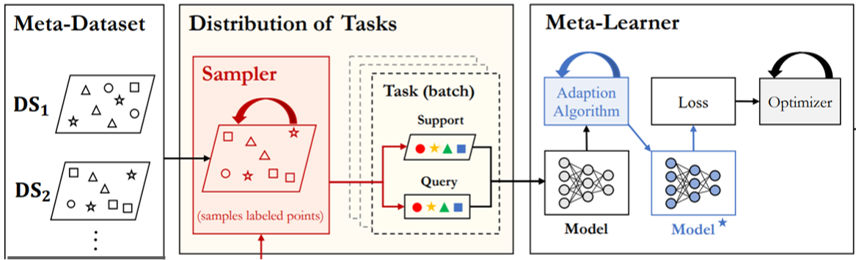}
    \caption{Overview of a typical meta-learning pipeline. We propose to enhance different steps like task sampling, task construction, mini-batching using known task-relations to improve the performance and robustness of meta-learning algorithms.}
    \label{fig:overview-pipeline}
\end{figure}

Fig. \ref{fig:overview-pipeline} shows the workflow of a typical meta-learning algorithm. The first step is to construct a meta-dataset which consists of various labelled datapoints. The sampler then samples from the meta-dataset (ususally randomly) to create tasks for the meta-training phase. The meta-training phase itself consists of two components. The support set of the meta-training task consist of labelled datapoints for supervised training while the query set consist of 'test datapoints'. During training, the meta-learner is optimized to learn from the support set tasks to give correct predictions (classification/regression) for the query set tasks. The meta-learner is optimized via the loss function which evaluates its performance over the query set. The performance of the trained meta-learner is evaluated during the meta-test phase where we judge how well and robustly it's able to learn from the support set tasks and apply its learnings to solve the query set tasks. Note that the meta-learner was not optimized over the support and query set of the meta-test phase.

Given the high-level overview of the meta-learning pipeline, we will now focus on two specific meta-learning algorithms namely Model-Agnostic Meta-Learning (MAML)  \cite{finn2017model} and Prototypical Networks (ProtoNets) \cite{snell2017prototypical, jindal2023classification}. The key idea of Model-Agnostic Meta-Learning (MAML) algorithm is to optimize model which can adapt to new task quickly. MAML provides a good initialization of a model’s parameters to achieve an optimal fast learning on a new task with only a small number of gradient steps while avoiding overfitting that may happen when using a small dataset. The second algorithm is Prototypical Networks (ProtoNets) which significantly differs from MAML design but performs the same task of few-shot classification. Here a classifier must generalize to new classes not seen in the training set, given only a small number of examples of each new class. Prototypical networks learn a metric space in which classification can be performed by computing distances to prototype representations of each class. Compared to recent approaches for few-shot learning, they reflect a simpler inductive bias that is beneficial in this limited-data regime, and achieve excellent results.

It's well established that meta-learning algorithms generalize well for tasks with low-data. However,  one of the biggest limitations to meta learning is the problem of negative transfer. Meta learning only works if the meta-training tasks are similar enough for the training to be relevant. If the meta training is too far off the mark, the model may actually perform worse than if it had never been trained at all. Hence, it is imperative to study the task distribution and if possible, augment the data with external sources of information. 

In this project, we aim to inform the task distribution of meta-train and meta-test tasks through an external source of information - namely the WordNet hierarchy. Specifically, we generate mini-batch of tasks with varying levels of task complexity and study the performance of MAML and ProtoNets for varying levels of meta-test task complexity. 

The paper is outlined a follows: Section \ref{relwork} gives details about the existing literature in this problem domain. In Section \ref{dataset}, we describe the dataset along with the approach to create the class relation graph and artificial dataset. In Section \ref{method}, we explain our approach to test the effect of task complexities on the performance of the meta-learning algorithm at test time. Section \ref{results} outlines the evaluation routines and presents the study results. Finally, we conclude our findings in Section \ref{conclude} followed by potential future work in Section \ref{future}.

\section {Related Work} \label{relwork}
Meta-learning involves learning new tasks quickly with a few training examples utilizing information from related tasks. However, the tasks may be very diverse, and the generalization across the entire task distribution may be ineffectual (if the test task distribution lies in some small clusters in the task distribution space) or even harmful to unrelated tasks (often referred to as negative transfer). Motivated by these observations, there have been many recent works on handling task heterogeneity in meta-learning \cite{yoon2019tapnet, vuorio2019multimodal, oreshkin2018tadam, li2019lgm, lee2018gradient, tadanet, yao2020automated, singh2018footwear}. These approaches can be divided into two broad categories, \cite{yoon2019tapnet, 10.1145/3308560.3316599, oreshkin2018tadam, li2019lgm, singh2019one} enforce task-specific representations instead of globally shared parameters but ignore the relationships between tasks, which may limit the model expressiveness and impair knowledge generalization. On the other hand, \cite{yao2020automated, yao2019hierarchically, rastogi2023exploring} learn task relationships from the training data but
ignore the inherent relationship expressed by external knowledge. \cite{tadanet,kumari2017parallelization} explores the use of externally available relationship between tasks (specifically the hierarchy of classes from which the task distribution is sampled) to learn a task embedding that characterizes task relationships and tailors task-specific parameters, resulting in a task-adaptive metric space for classification. All these methods show that incorporating either learned or external knowledge about the relation between tasks helps the meta-learner better generalize to unseen tasks \cite{rajan2023shaping}. Motivated by these works, we explore the benefits of using external knowledge about task and class relations to improve the efficiency and stability of different meta-learning algorithms.

\section{Dataset} \label{dataset}

We use the publicly available ImageNet \cite{russakovsky2015imagenet} dataset to construct a few-shot classification dataset for our work. We chose to create out own subsample of Imagenet classes because the current existing benchmarks like mini-ImageNet \cite{vinyals2016matching} have classes random selected which are all very distinct from one another and hence not suitable to demonstrate the issue of task quality that we study in this work. We first create a class relationship graph and then use clustering on the graph to sample classes such that is has a good mix of closely related classes and distinct classes as described below.

\subsection{Class relation graph}
\begin{figure}[ht]
    \centering
    \includegraphics{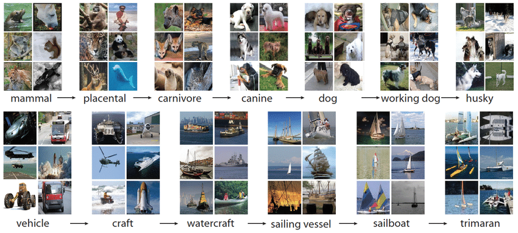}
    \caption{The figure shows the hierarchical nature of the classes on the ImageNet dataset where the hierarchy is defined by the hypernym relation in the WordNet graph}
    \label{fig:imagenet-hierarchy}
\end{figure}

WordNet \cite{miller1995wordnet} is one of the most popular thesauri for computational purposes in the NLP domain. WordNet contains all sorts of interesting relationships between words from over 200 languages. It can link words into semantic relations, including synonyms, and can categorize words into word hierarchies. We utilize the synset ids corresponding to the ImageNet classes and the hypernym relations to build a class-relation graph for all the 1000 classes Imagenet dataset. Fig. \ref{fig:imagenet-hierarchy} shows a few examples of the hierarchy defined by WordNet on the classes of the ImageNet dataset. We get a graph with 32,324 nodes (corresponding to different synsets) with 32,544 edges between them.

\subsection{Sampling classes}
The goal of our dataset is to have classes that are close to each other (like beagle and hound) and also classes that are distinct from each other (snakes and fruits). Hence given the class relation graph as defined above, we select nodes such that it has between 15 to 25 ImageNet classes and then sample 10 classes from the children of each such node. The classes in the same subtree are more closely related than the classes in different subtrees. For example, the class \texttt{moped} is more closelt related to \texttt{cab} in the same subtree \texttt{motor vehicle} than to other classes like \texttt{baboon} in a different subtree. Hence using this criteris, we sample 15 subtrees, 10 classes in each subtree and 100 images in each class to finally have a dataset with 150 classes and 15k images. 

\begin{figure}
    \centering
    \includegraphics[width=\linewidth]{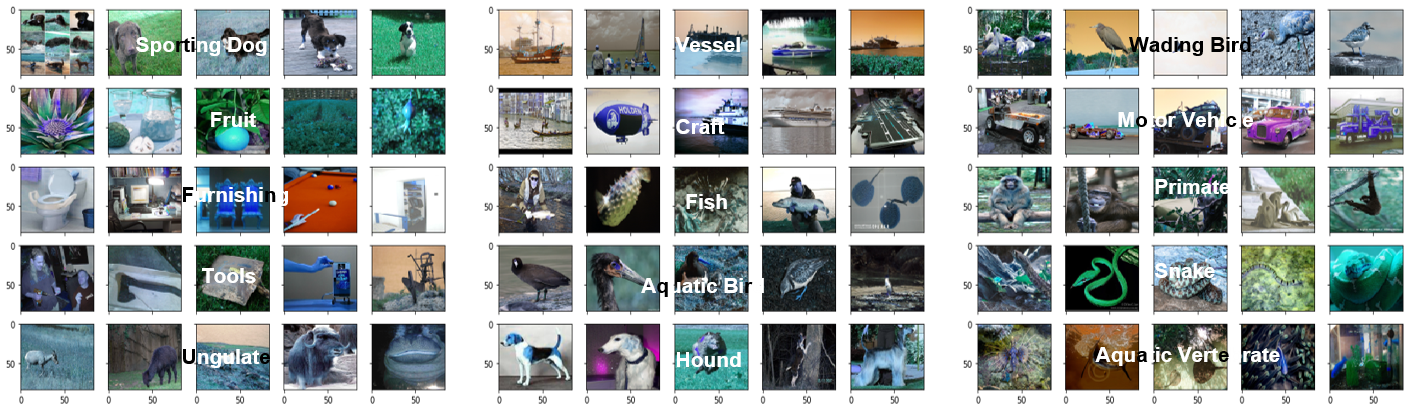}
    \caption{Overview of the dataset. Each row of 5 images represent images from classes in the corresponding cluster}
    \label{fig:dataset}
\end{figure}

Fig. \ref{fig:dataset} shows a sample of classes from each cluster in the dataset. As is evident, classes in the same cluster are similar and harder to classify into different classes.

\section{Methodology} \label{method}
In our work, we study the effects of complexities of tasks seen at meta-training time on the performance of meta-learning algorithms at test-time. 

\begin{figure}[ht]
    \centering
    \includegraphics[scale=0.7]{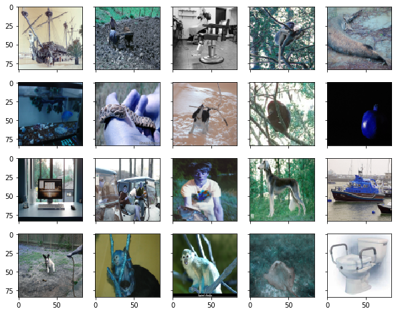}
    \caption{A batch of tasks for 5-way classification consisting of only \textbf{random tasks}}
    \label{fig:random_task_batch}
\end{figure}

We define a \textbf{random task} as a task consisting of classes randomly sampled from the pool of available classes such that the classes might be very distinct and easy for a ML model to differentiate between them. An example of randomly created tasks is shown in Fig. \ref{fig:random_task_batch}. As can be seen from the figure, the classes in a task are quite different like a television and a boat and hence it is very easy for a model to differentiate between them based on superficial signal only without developing an understanding of the image.

\begin{figure}[ht]
    \centering
    \includegraphics[scale=0.7]{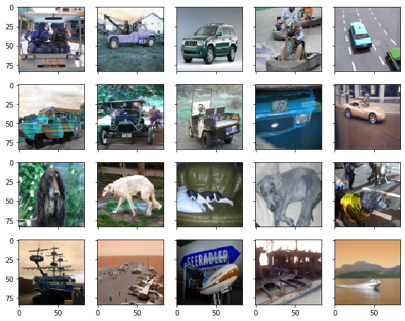}
    \caption{A batch of tasks for 5-way classification consisting of only \textbf{hard tasks}}
    \label{fig:hard_task_batch}
\end{figure}

We define a \textbf{hard task} as a task having classes which are similar to each other and hence hard for a ML model to distinguish between them. An example of a hard task is shown in Fig. \ref{fig:hard_task_batch}. As can be seen from the figure, the classes in the task are closely related like different species of dogs and hence a model needs to develop a good understanding of the images to be able to correctly distinguish between the classes.

\begin{figure}[ht]
    \centering
    \includegraphics[scale=0.6]{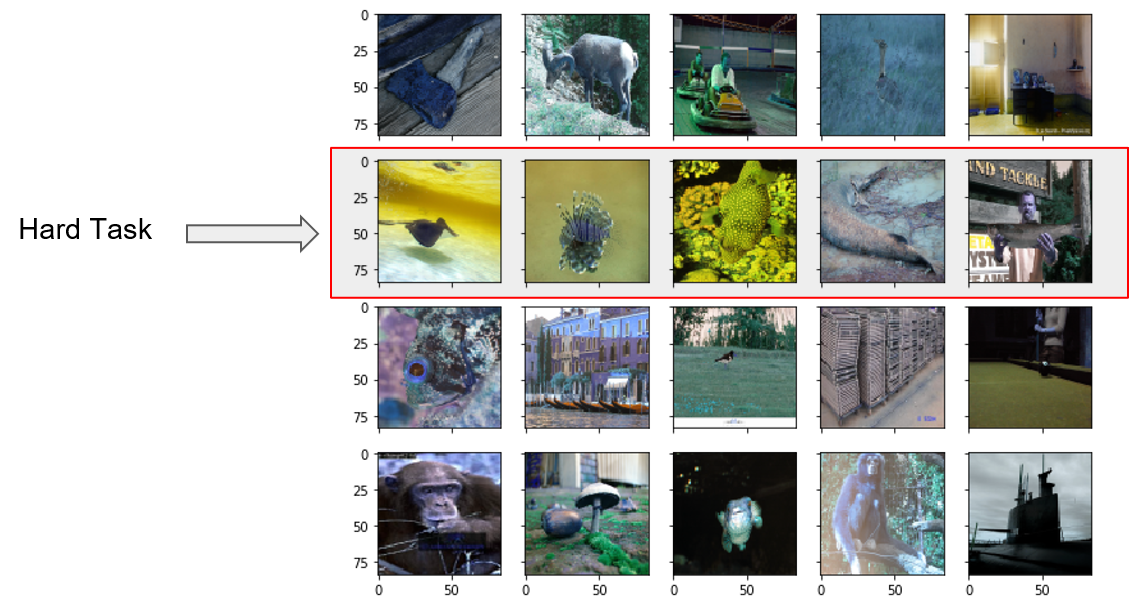}
    \caption{A batch of tasks for 5-way classification consisting of mixture of \textbf{hard tasks} and \textbf{random tasks}}
    \label{fig:mixed_task_batch}
\end{figure}

To study the effects of the task complexity on the meta-learner we devise three training regimes, namely:
\begin{enumerate}
    \item \textbf{Random task meta-training}: In this training regime, all tasks are sampled such that it consists of randomly selected classes from the 150 classes in the dataset and hence all the tasks seen by the meta-learner during meta-train time are \textbf{random tasks}. Fig. \ref{fig:random_task_batch} shows a batch of tasks encountered in this training regime.
     \item \textbf{Hard task meta-training}: In this training regime, all tasks are sampled such that the classes in the task belong to a randomly selected cluster out of the 15 clusters in the dataset and hence all the tasks seen by the meta-learner during meta-train time are \textbf{hard tasks}. Fig. \ref{fig:hard_task_batch} shows a batch of tasks encountered in this training regime.
    \item \textbf{Mixed task meta-training}: In this training regime, a \textbf{random task} is sampled with probability 0.5 and a \textbf{hard task} is sampled with probability 0.5. Hence the meta-learner sees \textbf{hard tasks} 50\% of time and \textbf{random tasks} other 50\% of the time during meta-train. Fig. \ref{fig:mixed_task_batch} shows a batch of tasks encountered in this training regime.
\end{enumerate}

To evaluate the performance of meta-learners with different test task distribution, we test the meta-learner with tasks drawn with different probabilities of seeing a \textbf{hard task}. We evaluate the performance of the meta-trainer trained using MAML and ProtoNets algorithms. We calculate the mean accuracy of the meta-learner on 1.6k tasks along with the 95\% confidence interval.

\section{Results and Analysis} \label{results}
\begin{figure}[ht]
    \centering
    \includegraphics[scale=0.8]{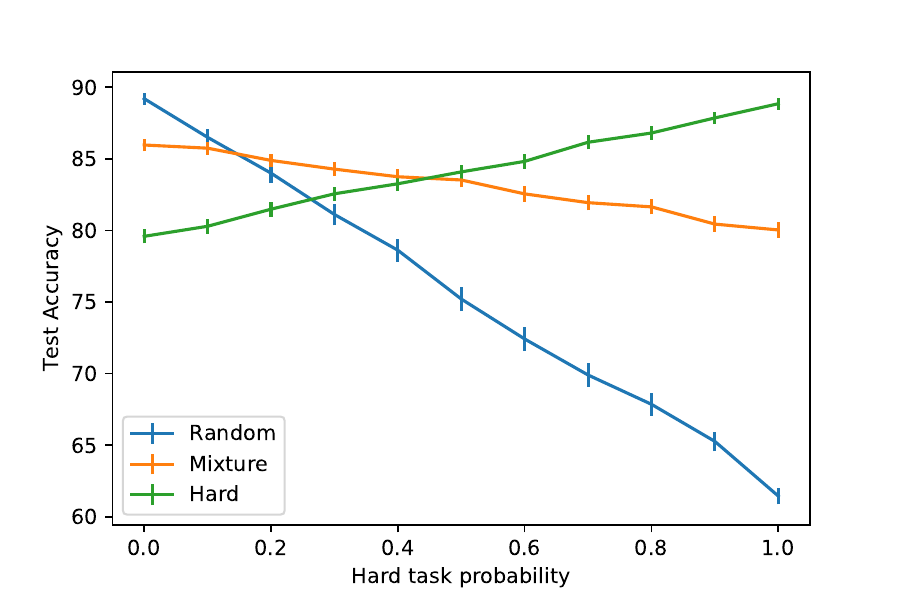}
    \caption{Plot of performance of the meta-learners trained using MAML in the three different training regimes on meta-test task distribution with varying \textbf{hard task} probability from 0 to 1}
    \label{fig:maml_perf}
\end{figure}

We use the TorchMeta library \cite{deleu2019torchmeta, 10.1007/978-981-10-8639-7_25} to train a 5-way 5-shot meta-learner for all out experiments.
Fig. \ref{fig:maml_perf} shows the performance of the meta-learners trained using MAML in the three different training regimes on test tasks distribution with varying \textbf{hard task} probability from 0 to 1. The figure shows that the performance of the meta-learner trained in the random task regime drops drastically as the probability of hard task increases in the meta-test phase. This is in accordance with our hypothesis that meta-learners only seeing random tasks in training latches on superficial features to distinguish between the classes and hence when faced with complex task in test time fails to generalize well. The performance of the meta-learner trained in hard task regime also suffers a small drop in performance when faced with random tasks. Our hypothesis is that the meta-learner have never seen distinct classes and hence does not have the ability to classify them properly. From the plot, we can observe that the meta-learner trained in the mixed task regime has a stable performance across all the test task distributions. This suggests that meta-learning algorithms benefit from training over a wide variety of tasks with different complexities so that the model can adapt well to any task at test-time. 

\begin{figure}[ht]
    \centering
    \includegraphics[scale=0.8]{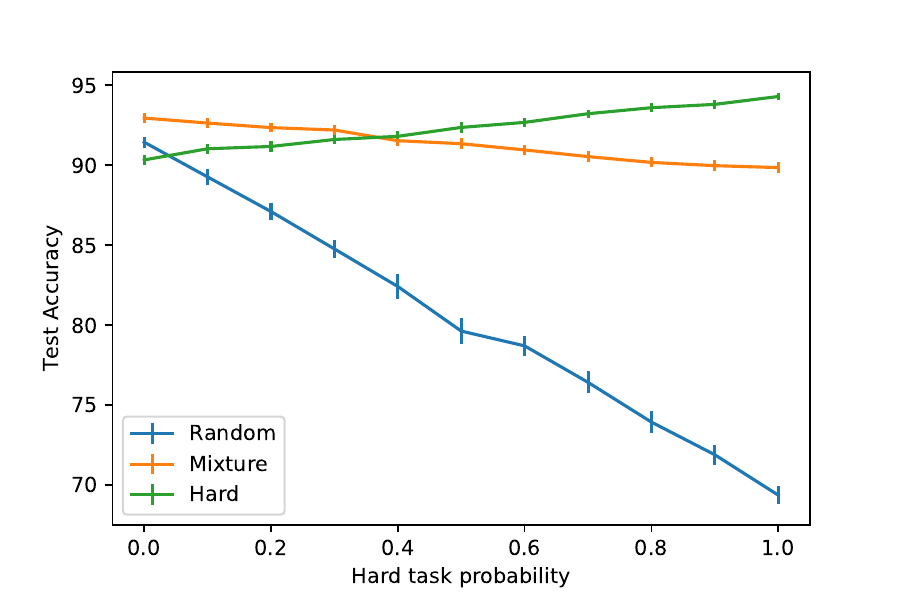}
    \caption{Plot of performance of the meta-learners trained using ProtoNet in the three different training regimes on meta-test task distribution with varying \textbf{hard task} probability from 0 to 1}
    \label{fig:proto_perf}
\end{figure}

Fig. \ref{fig:proto_perf} shows the performance of the meta-learners trained using ProtoNet in the three different training regimes on test tasks distribution with varying \textbf{hard task} probability from 0 to 1. We observe a similar characteristics as observed in the case of MAML above. Though in this case, the model trained only on hard tasks performs quite well on task of all difficulty levels and is similar to that in the mixed task regime. We surmise that this happens because ProtoNet learns efficient feature extractor for the images and training for hard tasks makes the representation more meaningful and hence can help in distinguishing between any two given classes.

\section{Conclusion} \label{conclude}

The huge drop in performance of the meta-learner trained in the random task regime when faced with increasingly hard meta-test tasks from both MAML and ProtoNet shows that random task creation is not the effective way to generate tasks for meta-training. This also suggests that the current benchmarks containing only distinct classes are not well equipped to gauge the performance of meta-learning algorithms because often meta-learning algorithms are employed in scenarios where the test tasks are quite varied - some of these test tasks can have distinct classes while some can consist of similar classes. For a meta-learning model to be robust and able to generalize well across a meta-test task distribution of varying complexity, it is imperative to focus on the meta-training task distribution and devise methods for the efficient training of the models. We conclude that having a uniform mix of random and hard task distribution during meta-training boosts the generalizability of the meta-learner. Finding the hard tasks is problem-specific. For example, in our problem setting, we were solving a classification problem and termed a task as a 'hard task' if all the classes for that task belong to the same cluster. These clusters were based on the WordNet hierarchy which helped us get more information about classes and their associated hierarchies. A hard task makes it difficult for the meta-learner to discriminate among classes. Similarly, this approach can be further expanded in other problem settings by first creating a bunch of hard tasks for the meta-learner followed by training on them. We believe, following this approach would benefit the learning process given the status-quo.

\section{Future Work} \label{future}

In the future, we plan on evaluating black-box algorithms in conjunction with more recent and advanced meta-learning algorithms over our proposed approach of effective mini-batching of tasks. We also plan to evaluate the performance of various data augmentation techniques for images like image flipping, rotation, shift as against our current approach of cluster-based sampling of tasks. Lastly, we wish to explore the relationship between meta-train and meta-test tasks and leverage that information to select meta-train tasks which are tightly correlated to the meta-test tasks. 
\printbibliography

@inproceedings{tadanet,
author = {Suo, Qiuling and Chou, Jingyuan and Zhong, Weida and Zhang, Aidong},
title = {TAdaNet: Task-Adaptive Network for Graph-Enriched Meta-Learning},
year = {2020},
isbn = {9781450379984},
publisher = {Association for Computing Machinery},
address = {New York, NY, USA},
url = {https://doi.org/10.1145/3394486.3403230},
doi = {10.1145/3394486.3403230},
booktitle = {Proceedings of the 26th ACM SIGKDD International Conference on Knowledge Discovery \&amp; Data Mining},
pages = {1789–1799},
numpages = {11},
location = {Virtual Event, CA, USA},
series = {KDD \'20}
}

@article{lee2018gradient,
  title={Gradient-based meta-learning with learned layerwise metric and subspace},
  author={Lee, Yoonho and Choi, Seungjin},
  journal={arXiv preprint arXiv:1801.05558},
  year={2018}
}

@article{li2019lgm,
  title={Lgm-net: Learning to generate matching networks for few-shot learning},
  author={Li, Huaiyu and Dong, Weiming and Mei, Xing and Ma, Chongyang and Huang, Feiyue and Hu, Bao-Gang},
  journal={arXiv preprint arXiv:1905.06331},
  year={2019}
}

@inproceedings{oreshkin2018tadam,
  title={Tadam: Task dependent adaptive metric for improved few-shot learning},
  author={Oreshkin, Boris and L{\'o}pez, Pau Rodr{\'i}guez and Lacoste, Alexandre},
  booktitle={Advances in Neural Information Processing Systems},
  pages={721--731},
  year={2018}
}

@inproceedings{vuorio2019multimodal,
  title={Multimodal Model-Agnostic Meta-Learning via Task-Aware Modulation},
  author={Vuorio, Risto and Sun, Shao-Hua and Hu, Hexiang and Lim, Joseph J},
  booktitle={Advances in Neural Information Processing Systems},
  pages={1--12},
  year={2019}
}

@article{yoon2019tapnet,
  title={Tapnet: Neural network augmented with task-adaptive projection for few-shot learning},
  author={Yoon, Sung Whan and Seo, Jun and Moon, Jaekyun},
  journal={arXiv preprint arXiv:1905.06549},
  year={2019}
}

@article{yao2019hierarchically,
  title={Hierarchically structured meta-learning},
  author={Yao, Huaxiu and Wei, Ying and Huang, Junzhou and Li, Zhenhui},
  journal={arXiv preprint arXiv:1905.05301},
  year={2019}
}

@article{yao2020automated,
  title={Automated relational meta-learning},
  author={Yao, Huaxiu and Wu, Xian and Tao, Zhiqiang and Li, Yaliang and Ding, Bolin and Li, Ruirui and Li, Zhenhui},
  journal={arXiv preprint arXiv:2001.00745},
  year={2020}
}

@article{finn2017model,
  title={Model-agnostic meta-learning for fast adaptation of deep networks},
  author={Finn, Chelsea and Abbeel, Pieter and Levine, Sergey},
  journal={arXiv preprint arXiv:1703.03400},
  year={2017}
}

@inproceedings{snell2017prototypical,
  title={Prototypical networks for few-shot learning},
  author={Snell, Jake and Swersky, Kevin and Zemel, Richard},
  booktitle={Advances in neural information processing systems},
  pages={4077--4087},
  year={2017}
}

@article{miller1995wordnet,
  title={WordNet: a lexical database for English},
  author={Miller, George A},
  journal={Communications of the ACM},
  volume={38},
  number={11},
  pages={39--41},
  year={1995},
  publisher={ACM New York, NY, USA}
}

@inproceedings{vinyals2016matching,
  title={Matching networks for one shot learning},
  author={Vinyals, Oriol and Blundell, Charles and Lillicrap, Timothy and Wierstra, Daan and others},
  booktitle={Advances in neural information processing systems},
  pages={3630--3638},
  year={2016}
}

@article{russakovsky2015imagenet,
  title={Imagenet large scale visual recognition challenge},
  author={Russakovsky, Olga and Deng, Jia and Su, Hao and Krause, Jonathan and Satheesh, Sanjeev and Ma, Sean and Huang, Zhiheng and Karpathy, Andrej and Khosla, Aditya and Bernstein, Michael and others},
  journal={International journal of computer vision},
  volume={115},
  number={3},
  pages={211--252},
  year={2015},
  publisher={Springer}
}

@misc{deleu2019torchmeta,
  title={{Torchmeta: A Meta-Learning library for PyTorch}},
  author={Deleu, Tristan and W\"urfl, Tobias and Samiei, Mandana and Cohen, Joseph Paul and Bengio, Yoshua},
  year={2019},
  url={https://arxiv.org/abs/1909.06576},
  note={Available at: https://github.com/tristandeleu/pytorch-meta}
}

@inproceedings{10.1145/3308560.3316599,
author = {Mohammed Abdulla, G and Singh, Shreya and Borar, Sumit},
title = {Shop Your Right Size: A System for Recommending Sizes for Fashion Products},
year = {2019},
isbn = {9781450366755},
publisher = {Association for Computing Machinery},
address = {New York, NY, USA},
url = {https://doi.org/10.1145/3308560.3316599},
doi = {10.1145/3308560.3316599},
booktitle = {Companion Proceedings of The 2019 World Wide Web Conference},
pages = {327–334},
numpages = {8},
location = {San Francisco, USA},
series = {WWW \'19}
}

@article{singh2019one,
  title={One embedding to do them all},
  author={Singh, Loveperteek and Singh, Shreya and Arora, Sagar and Borar, Sumit},
  journal={arXiv preprint arXiv:1906.12120},
  year={2019}
}

@article{singh2018footwear,
  title={Footwear Size Recommendation System},
  author={Singh, Shreya and Abdulla, G Mohammed and Borar, Sumit and Arora, Sagar},
  journal={arXiv preprint arXiv:1806.11423},
  year={2018}
}

@article{kumari2017parallelization,
  title={Parallelization of alphabeta pruning algorithm for enhancing the two player games},
  author={Kumari, SDGV Akanksha and Singh, Shreya},
  journal={Int. J. Advances Electronics Comput. Sci},
  volume={4},
  pages={74--81},
  year={2017}
}

@InProceedings{10.1007/978-981-10-8639-7_25,
author="Sebastian, Abraham Gerard
and Singh, Shreya
and Manikanta, P. B. T.
and Ashwin, T. S.
and Reddy, G. Ram Mohana",
editor="Sa, Pankaj Kumar
and Bakshi, Sambit
and Hatzilygeroudis, Ioannis K.
and Sahoo, Manmath Narayan",
title="Multimodal Group Activity State Detection for Classroom Response System Using Convolutional Neural Networks",
booktitle="Recent Findings in Intelligent Computing Techniques ",
year="2019",
publisher="Springer Singapore",
address="Singapore",
pages="245--251",
isbn="978-981-10-8639-7"
}

@article{jindal2023classification,
  title={Classification for everyone: Building geography agnostic models for fairer recognition},
  author={Jindal, Akshat and Singh, Shreya and Gadgil, Soham},
  journal={arXiv preprint arXiv:2312.02957},
  year={2023}
}

@article{rastogi2023exploring,
  title={Exploring Graph Based Approaches for Author Name Disambiguation},
  author={Rastogi, Chetanya and Agarwal, Prabhat and Singh, Shreya},
  journal={arXiv preprint arXiv:2312.08388},
  year={2023}
}

@article{rajan2023shaping,
  title={Shaping Political Discourse using multi-source News Summarization},
  author={Rajan, Charles and Asnani, Nishit and Singh, Shreya},
  journal={arXiv preprint arXiv:2312.11703},
  year={2023}
}
\end{document}